\documentclass[conference]{IEEEtran}
\IEEEoverridecommandlockouts
\usepackage{cite}
\usepackage{amsmath,amssymb,amsfonts}
\usepackage{algorithmic}
\usepackage{graphicx}
\usepackage{tikz}
\usepackage{textcomp}
\usepackage{hyperref}
\usepackage{ipdps26repro}
\usepackage{tagging}
\usepackage[caption=false,font=normalsize,labelfont=sf,textfont=sf]{subfig}
\usepackage{xcolor}

\usepackage{booktabs}
\usepackage{multirow}
\hypersetup{
    colorlinks=true,
    linkcolor=blue,
    citecolor=blue,
    urlcolor=blue
}
\usepackage[linesnumbered,ruled,vlined]{algorithm2e}
\def\BibTeX{{\rm B\kern-.05em{\sc i\kern-.025em b}\kern-.08em
    T\kern-.1667em\lower.7ex\hbox{E}\kern-.125emX}}
\newcommand{\circled}[1]{%
  \tikz[baseline=(char.base)]{%
    \node[shape=circle, draw, inner sep=0.5pt] (char) {#1};%
  }%
}
\usepackage{subcaption}
\usepackage{fancyhdr}

\begin{document}

\title{FedACT: Concurrent Federated Intelligence across Heterogeneous Data Sources
\\
} 

\author{
    \IEEEauthorblockN{Md Sirajul Islam\textsuperscript{1}, Isabelle G Chapman\textsuperscript{1}, N I Md Ashafuddula\textsuperscript{1}, Xu Yuan\textsuperscript{2}, Li Chen\textsuperscript{1}, \\Nian-Feng Tzeng\textsuperscript{1}, and Klara Nahrstedt\textsuperscript{3}}
    \IEEEauthorblockA{\textsuperscript{1}School of Computing and Informatics, University of Louisiana at Lafayette, USA\\
    \textsuperscript{2}Department of Computer and Information Sciences, University of Delaware, USA \\
    \textsuperscript{3}Department of Computer Science, University of Illinois at Urbana-Champaign, USA}
    \thanks{The research was supported by the National Science Foundation under grant OIA-2327452.}
    \thanks{Corresponding author: Li Chen. Email: li.chen@louisiana.edu}
}
\maketitle

\begin{abstract}
Federated Learning (FL) enables collaborative intelligence across decentralized data source devices in a privacy-preserving way. While substantial research attention has been drawn to optimizing the learning process for an individual task, real-world applications increasingly require multiple machine learning tasks simultaneously training their models across a shared pool of devices. Naively applying single-FL optimization techniques in multi-FL systems results in suboptimal system performance, particularly due to device heterogeneity and resource inefficiency. To address such a critical open challenge, we introduce {\em FedACT}, a novel resource heterogeneity-aware device scheduling approach designed to efficiently schedule heterogeneous devices across multiple concurrent FL jobs, with the goal of minimizing their average job completion time (JCT). {\em FedACT} dynamically assigns devices to FL jobs based on an alignment scoring mechanism that evaluates the compatibility between available resources of devices and resource demands of jobs. Additionally, it incorporates participation fairness to ensure balanced contributions from devices across jobs, further enhancing the accuracy levels of learned global models. An optimal scheduling plan is formulated in {\em FedACT} by prioritizing devices with higher alignment scores, while ensuring fair participation across jobs. To evaluate the effectiveness of the proposed scheduling algorithm, we carried out comprehensive experiments using diverse FL jobs and benchmark datasets. Experimental results demonstrate that {\em FedACT} reduces the average JCT by up to 8.3\(\times\) and improves model accuracy by up to 44.5\%, compared to the state-of-the-art baselines.
\end{abstract}


\section{Introduction}
\label{sec:introduction}
The rapid proliferation of Internet of Things (IoT) and mobile devices equipped with advanced sensing capabilities has led to the generation of vast amounts of decentralized data, which can be leveraged to develop machine learning models for a wide range of intelligent applications. Due to the sensitive nature of user data and the constraints of privacy regulations and data protection laws\cite{gdpr}, traditional learning methods that require data gathering into a central datacenter become infeasible. Federated Learning (FL) \cite{mcmahan2017communication,li2020federated} enables collaborative model training across distributed edge devices without requiring the transfer of raw data. As a privacy-preserving paradigm, FL has been widely adopted in various edge-intelligent applications, including home automation\cite{zhou2020falcon}, health monitoring\cite{ouyang2021clusterfl}, and typing assistance\cite{hard2018federated}.

\begin{figure}[t]  
\centering
\includegraphics[width=0.95\columnwidth]{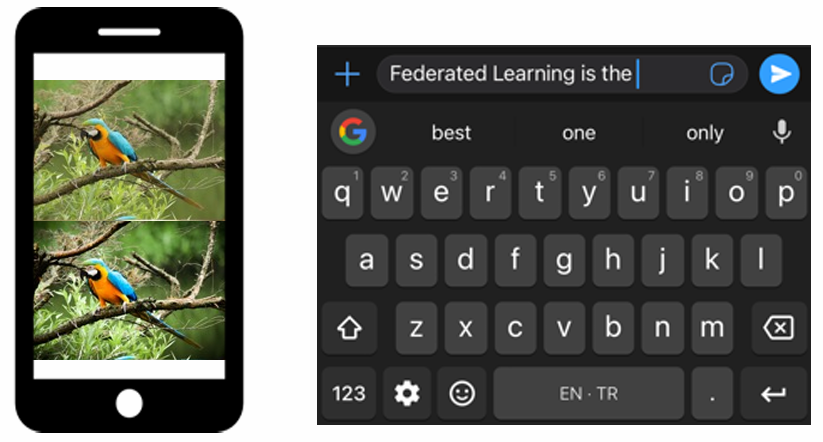}
\caption{An example of Multi-Job Federated Learning: smartphones simultaneously improve next-word prediction for keyboards while enhancing image quality.}
\label{example-intro}
\vspace{-15pt}
\end{figure}

Despite its promise, the real-world deployment of FL is still challenging due to the presence of non-independent and identically distributed (non-IID) data across devices (data heterogeneity), variations in device hardware capabilities (system heterogeneity), and the coexistence of multiple machine learning models to be simultaneously trained. To improve the efficiency and overall performance of FL systems, existing efforts have investigated different aspects, including device selection \cite{lai2021oort,li2022pyramidfl,javaherian2024fedfair}, communication efficiency \cite{islam2024fedclust,wu2022communication,hamer2020fedboost}, personalization \cite{li2021ditto, luo2022pgfed, zhang2023fedala}, and issues related to scalability and fast training \cite{xie2019asynchronous, nguyen2022federated, guo2024fedcore, Islam2025SEAFL}. However, these studies focus on the federated learning of an individual model, without being aware of the coexistence of multiple models to be learned for different tasks in real-world applications. For example, decentralized smartphones may simultaneously learn a language model for next-word prediction and an image recommendation model to suggest photos likely to be shared, as illustrated in Fig. \ref{example-intro}. Similarly, a chat application may need both speech recognition and response text generation models to function concurrently.

To accommodate the simultaneous training of multiple models in vanilla FL, a straightforward way is sequential dispatching, with each model training job launched and completed one after another using available devices. However, it remains an open challenge to identify an optimal schedule given the heterogeneity of both data distribution and device capability. Moreover, the devices unselected for the current model training job could have been better utilized to contribute to the progress of another model learning, speeding up the completion of the entire set of jobs. Towards enhancing the efficiency of multi-job FL, MJFL \cite{zhou2022efficient} relies on a Bayesian optimization-based scheduling approach to select a subset of devices for each job based on their computational and communication capabilities, while FedAST \cite{askin2024fedast} introduced a buffered asynchronous FL aggregation mechanism to mitigate the straggler issue during the simultaneous training of multiple jobs. Despite their advancements over conventional single-job FL, these multi-job FL attempts overlook the variation in device performance across different models (e.g., large language models versus small image classifiers). Thus, they fail to assign jobs to devices with optimal resource demand matching, inevitably leading to inefficient resource utilization and prolonged overall model training time.




In this paper, we propose {\em FedACT}, a novel resource heterogeneity-aware scheduling strategy for multi-job FL systems, designed to efficiently learn multiple models across heterogeneous edge devices. The key idea is to dynamically assign heterogeneous devices to diverse jobs based on optimal device-job matching through an alignment scoring mechanism. This score quantifies the compatibility between the available resources of a device and the resource demands of a job. Additionally, to ensure sufficient contribution from representative devices across concurrent jobs, {\em FedACT} incorporates participation fairness into the alignment score. The fairness module adjusts per-device alignment score based on its participation frequency, discouraging the over-representation of data samples from resource-preferred devices that are frequently selected while promoting underrepresented ones. During the multi-job FL training process, {\em FedACT} dynamically updates alignment scores for each device-job pair and generates a scheduling plan that prioritizes devices with higher alignment scores, striking for an optimal balance between training efficiency and global model accuracy. With a comprehensive consideration and handling of intricate heterogeneity factors, the proposed framework manages to identify optimal schedules of devices for concurrent FL jobs across time, improving resource efficiency and minimizing the overall job completion time for concurrent model training.

To validate the effectiveness of {\em FedACT}, we conduct extensive experiments on various image classification tasks using five benchmark datasets (CIFAR-10 \cite{krizhevsky2009learning}, MNIST \cite{lecun1998gradient}, EMNIST-Letters\cite{cohen2017emnist}, EMNIST-Digits\cite{cohen2017emnist}, and Fashion-MNIST \cite{xiao2017fashion}) and five different models, in comparison with existing multi-job FL schedulers. More specifically, our experiments involve multiple concurrent FL jobs divided into two distinct groups, using models such as LeNet, CNN, VGG, ResNet-18, and AlexNet. We analyze the wall-clock training time required to reach a target accuracy and the average job completion time (JCT) for each job group under both IID and non-IID settings. Experimental results demonstrate that {\em FedACT} significantly improves global model accuracy by up to $\sim$44\% and reduces the average job completion time by up to 8.3\(\times\) compared to existing baselines. 

The rest of this paper is presented as follows. The background and motivation of this work are introduced in Section~\ref{sec:motivation}. Section~\ref{sec:design} presents the design of our proposed {\em FedACT} framework. We evaluate {\em FedACT} in Section~\ref{sec:experiment} and compare its performance with state-of-the-art counterparts. Section~\ref{sec:relatedwork}  describes related work, followed by Section~\ref{sec:end} with a concluding remark.

\section{Background and Motivation}
\label{sec:motivation}

\begin{table*}[t]
\caption{Notations and Descriptions}
\label{tab:notations}
\centering
\begin{tabular}{cc}
\toprule
Notation & Definition \\
\midrule
$\mathcal{K}$; $|\mathcal{K}|$ & Set of all devices; the total number of devices \\
$M$; $T$; $m$ & The total number of jobs; total training time; job index \\
$D_k^m$; $d_k^m$; $|D_k^m|$ & The local dataset on Device $k$ for Job $m$; batch size; size of $D_k^m$ \\
$D^m$; $|D^m|$ & Global dataset of Job $m$; size of $D^m$ \\
$F_k^m(\boldsymbol{w})$; $F^m(\boldsymbol{w})$ & Local loss function of Device $k$ for Job $m$; global loss function of Job $m$ \\
$\boldsymbol{w}_k^{m,r}(j)$ & Local model of Job $m$ in Device $k$ in the $j$-th local update of Round $r$ \\
$R_m$; $l_m$ & The maximum training rounds for Job $m$; the targeted final loss for Job $m$ \\
$\tau_m$; $C_m$ & The number of local training epochs of Job $m$; the fraction of the total devices $|\mathcal{K}|$ assigned to Job $m$  \\
$\alpha; \beta$ & The resource alignment weight; the participation fairness weight\\
$\boldsymbol{s}_m$ & The participation count vector associated with Job $m$ \\
$s_{k,m}^r$ & The participation frequency of Device $k$ for Job $m$ at Round $r$\\
$\mathcal{V}_m^r$ & The set of devices assigned to Job $m$ at Round $r$ \\
${p}_k^r$ & The weight assigned to Device $k$ during aggregation in round $r$\\
$\mathcal{V}_o$; $\mathcal{V}_o^r$ & A set of occupied devices; the set of occupied devices in Round $r$ \\
\bottomrule
\end{tabular}
\label{notation}
\vspace{-10pt}
\end{table*}

Federated learning has been recognized as a promising privacy-preserving solution to enabling the collaborative learning of a global model without exposing decentralized raw data \cite{mcmahan2017communication, li2020}. In classical FL training \cite{mcmahan2017communication, karimireddy2020scaffold, li20201federated}, the server broadcasts the current global model to the selected participating devices. Each device performs local training using its private data and then transmits model updates back to the server. The server aggregates these updates to refine the global model, which is subsequently used in the next training round. This iterative process is repeated until the global model achieves the target accuracy or a predefined number of communication rounds is completed. More specifically, the learning goal \cite{mcmahan2017communication} is to minimize the following objective function:
\setlength{\abovedisplayskip}{5pt} 
\begin{equation}
\min_{w} F(w) \overset{\scriptscriptstyle\Delta}{=} \sum_{k=1}^{K}\frac{n_k}{N}F_k(w)
\end{equation}
Here, $K$ is the total number of participating devices and each device $k$ has a local dataset $\mathcal{D}_k$. The number of local samples on device $k$ is denoted by $n_k =|\mathcal{D}_k|$, and the total number of samples across all devices is $N = \sum_{k=1}^{K} n_k$. Each device's local objective function is defined as the empirical loss computed over its local dataset $\mathcal{D}_k$, i.e., $F_k(w)=\frac{1}{n_k}\sum_{j_k=1}^{n_k}f_{j_k} (w; x_{j_k},y_{j_k})$.

In real-world scenarios, concurrent FL jobs (such as next-word prediction, object detection, and speech recognition) are increasingly witnessed, posing a critical open challenge for {\em fast and efficient multi-job federated learning in heterogeneous environments}. Training these models in parallel can maximally utilize available device resources, but it is unknown how to strategically assign devices across jobs throughout the training process to ensure system efficiency and learning outcome, especially given the prevailing existence of heterogeneity in both device resources and data distributions.



A straightforward solution is to train each job independently using the standard FL procedure, such as FedAvg \cite{mcmahan2017communication}, by directly applying scheduling strategies designed for single-job FL scenarios. Although this approach enables parallel execution, it often leads to inefficient utilization of available devices, resulting in suboptimal system performance. Furthermore, directly extending conventional single-job scheduling methods to multi-job FL fails to jointly optimize both training efficiency and model accuracy. 

In addition, the participation frequency of devices across jobs significantly influences both convergence speed and model performance. For example, repeatedly assigning specific powerful devices to a job may exclude valuable training samples from other devices, potentially limiting the model’s generalization ability. To accelerate the training of multiple jobs while maintaining high standards of model accuracy,  it is imperative to design a scheduling strategy that accounts for the intricate factors and addresses the open challenges in multi-job FL systems. As such, we are motivated to design {\em FedACT}, an efficient multi-job device scheduling solution, to be presented as follows.


\begin{figure*}
\centering
{\includegraphics[height=7cm, width=0.95\textwidth]{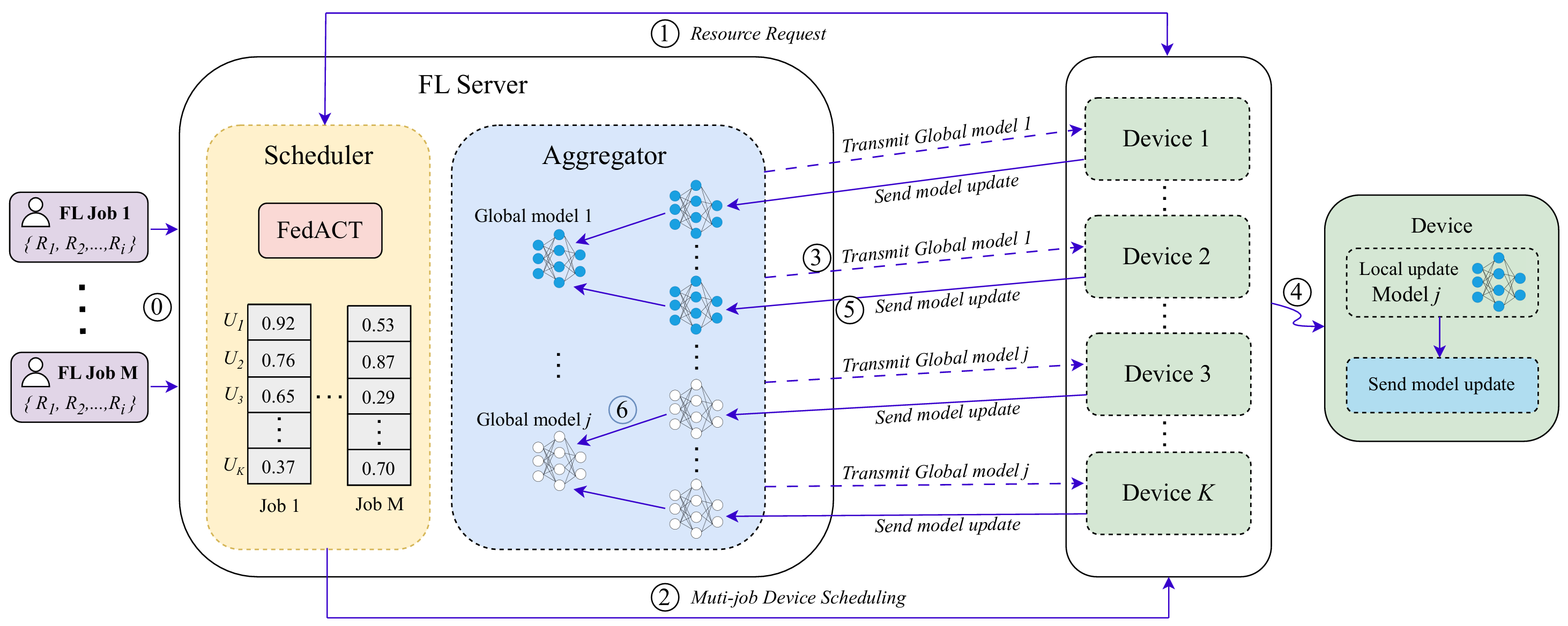}}
\caption{An overview of the training procedure within the {\em FedACT} framework.}
\label{overview}
\vspace{-13pt}
\end{figure*}
\section{System Design}
\label{sec:design}
In this section, we first present the problem statement, followed by the development of our proposed scheduling algorithm for the simultaneous training of multiple FL jobs. 

\subsection{Problem Statement}
We consider a multi-job FL setting with a set of \textit{K} devices, represented by $\mathcal{K}$, which collaboratively participate in the simultaneous training of multiple machine learning models. The training of each model is denoted as a job, which involves multiple local computation tasks launched on participating devices across multiple training rounds. The set of \textit{M} concurrent jobs is denoted by $\mathcal{M}$. Without loss of generality, each device $\textit{k} \in \mathcal{K}$ is assumed to hold local datasets for each job $\textit{m} \in \mathcal{M}$, denoted as $\mathcal{D}_k^m = \{x_{k,d}^m \in \mathbb{R}^{n_m}, y_{k,d}^m \in \mathbb{R}\}_{d=1}^{D_k^m}$, where $D_k^m = |\mathcal{D}_k^m|$ indicates the number of samples, $x_{k,d}^m$ refers to the \textit{d}-th input vector of dimension \textit{$n_m$}, and $y_{k,d}^m$ represents the labeled output of $x_{k,d}^m$. The complete dataset for Job \textit{m} is represented as $\mathcal{D}^m = \cup_{k \in \mathcal{K}} \mathcal{D}_k^m \text{ with the total size of } D^m = \sum_{k \in \mathcal{K}} D_k^m$. The primary goal of multi-job FL is to collaboratively learn the model parameters $\{w^m\}$ for each job $\textit{m} \in \mathcal{M}$ by leveraging the decentralized datasets distributed across participating devices. This global optimization problem in multi-job FL can be formulated as follows:
\begin{equation}
\min_{W} \sum_{m=1}^M \mathbb{L}_m, with \ \mathbb{L}_m = \sum_{k=1}^K \frac{D_k^m}{D^m} F_k^m(w^m)
\label{formula2}
\end{equation}
where $\mathbb{L}_m$ denotes the overall loss for Job \textit{m}, and the local loss at device \textit{k} for Job \textit{m} is defined as $F_k^m(w^m) = \frac{1}{D_k^m} \sum_{\{x_{k,d}^m, y_{k,d}^m\} \in \mathcal{D}_k^m} f^m(w^m; x_{k,d}^m, y_{k,d}^m)$. Here, $\boldsymbol{W} :\equiv \{w^1, w^2, ..., w^M\}$ represents the collection of model parameters for all jobs, and $f^m(w^m; x_{k,d}^m, y_{k,d}^m)$ computes the loss of model $w^m$ on the data sample $\{x_{k,d}^m, y_{k,d}^m\}$. For clarity, we present a list of key notations frequently used throughout this paper in Table \ref{notation}.

\subsection{FedACT Overview}

To solve the problem outlined in Eq.~(\ref{formula2}), the FL server needs to continuously assign devices across multiple jobs and update the global models iteratively until each job either converges or reaches the desired performance metrics, such as accuracy or loss. 

Fig.~\ref{overview} provides an overview of the training workflow within the {\em FedACT} framework. Before training starts, FL job managers submit their jobs to the FL server, specifying device requirements and associated resource demands for each resource type in Step \circled{0}. The server begins by initializing a global model for each job, which can be done either randomly or using pre-trained weights obtained from public datasets. {\em FedACT} queries the available devices to gather their current resource status in Step \circled{1}. Then, in Step \circled{2}, {\em FedACT} calculates an alignment score for each device-to-job pair (see Section~\ref{sec:scoring}) and generates a job-specific scheduling plan using the proposed scheduling strategy. This plan determines which devices will participate in local training for the current job. It is important to note that the {\em FedACT} dynamically produces a separate plan for each job during the training of multiple jobs, following an online approach, while the scheduling processes for different jobs occur concurrently. {\em FedACT} maintains the alignment scores and participation frequency counts for devices associated with each job. In Step \circled{3}, the server broadcasts the latest global model for the current job to the selected devices. Each device then updates the model based on its local data in Step \circled{4}. Subsequently, each device sends the updated model to the server after completing its local training in Step \circled{5}. Finally, in Step \circled{6}, the server aggregates the received updates from the scheduled devices to generate a new global model. This sequence of Steps \circled{1} to \circled{6}, referred to as a round, is executed iteratively for each job until the corresponding global model satisfies the target performance criteria, such as accuracy, convergence, or loss. Multiple jobs are executed asynchronously in parallel, with each device being assigned to only one job per round. Additionally, we assume that all jobs are treated with equal priority. Note that Steps  \circled{3} through \circled{6} follow conventional FL protocols between individual FL jobs and devices, while {\em FedACT}'s primary contribution lies in optimizing the job-to-device assigning phase, denoted by Step \circled{2}.
We assume that devices periodically report their current resource status to the FL server. The server only receives this information along with local model updates. Our framework ensures that sensitive data never leaves the device, and only the model updates and resource information are shared with the server. In addition, privacy-enhancing techniques such as homomorphic encryption \cite{paillier1999public}, secure aggregation \cite{bonawitz2016practical}, and differential privacy \cite{dwork2008differential} can be incorporated to further safeguard sensitive user information.

\subsection{Alignment Scoring}
\label{sec:scoring}
This section introduces the alignment scoring mechanism, designed to assess the compatibility between devices and jobs by jointly considering their resource capabilities and historical participation frequency. In order to measure the alignment score for each device-to-job pair throughout the training process, we exploit the scoring mechanism accounting for both resource alignment and participation fairness, defined as follows. 
\begin{equation}
Score^r_{k,m}(\mathcal{V}^r_m) = \alpha * \mathcal{R}^r_{k,m}(\mathcal{V}^r_m) + \beta * \mathcal{F}^r_{k,m}(\mathcal{V}^r_m)
\label{formula3}
\end{equation}
Here, $\alpha$ and $\beta$ represent the weights for resource alignment and participation fairness, respectively. $ \mathcal{R}^r_{k,m}$ denotes the resource alignment score between device \textit{k} and job \textit{m} in round \textit{r} with the set of available jobs, and $\mathcal{F}^r_{k,m}$ is the corresponding participation fairness score. {\em FedACT} schedules devices to jobs based on their alignment scores and dynamically updates these scores throughout the training process. In our experiments, the values of $\alpha$ and $\beta$ are empirically determined based on insights from previous runs and fine-tuned over short training epochs. A higher value of $\alpha$ accelerates training, while $\beta$ primarily contributes to improving model accuracy. We normalize the alignment scores for each device-to-job pair to the range [0, 1]. The pseudocode of {\em FedACT} is shown in Algorithm~\ref{algo1}. The following subsections elaborate on the two key components.

{\bf Resource Alignment.}
The score of resource alignment is used for matching devices to jobs based on the available resources on the devices and the resource requirements of each job. The objective is to assign devices such that those better equipped for a specific job should be prioritized for that job to ensure efficient resource utilization. We assume that job managers will specify the minimum resource requirements for their computation tasks upon submission, while devices periodically update the server with information about their available resources. For an available device, the scheduler first identifies a set of jobs whose minimum resource requirements are satisfied by that device. For each job in this set, the {\em FedACT} algorithm calculates a resource alignment score for the device. The alignment score is a weighted dot product between the device’s available resources and the job's minimum resource demands. The device with the highest alignment scores is then selected and assigned to that job. This process repeats until the required number of devices for each job is fulfilled.

We consider the resource requirements of jobs across three key types of resources: computational power, memory, and network bandwidth. When calculating the dot product, {\em FedACT} normalizes both the task's resource requirements and the device's available resources relative to the device's total capacity. This ensures that the numerical range of a device's resources and tasks's demands do not impact the alignment score. All the resources are given equal weight in our current setting. We randomly generate the memory capacity and network bandwidth within representative ranges for the available devices. To simulate the heterogeneous computational capabilities of devices, we model the execution time of each job using the conventional shifted exponential distribution \cite{shi2020joint,lee2017speeding}, expressed as follows: 
\begin{equation}
P[t^k_m<t] = 
\begin{cases}
1 - e^{-\frac{\mu_k}{\tau_m D^m_k}(t-\tau_m \alpha_k D^m_k)}, & t \geq \tau_m \alpha_k D^m_k \\
0, & \text{otherwise}
\end{cases}
\label{formula4}
\end{equation}
where $t^k_m$ denotes the execution time of job m on device \textit{k}. The parameters $\alpha_k > 0 \text{ and } \mu_k > 0$ indicate the maximum value and the fluctuation of the computational and communication capabilities of device \textit{k}, which are combined into one quantity. 
Eq.~(\ref{formula4}) ensures a minimum execution time proportional to both the job’s computational demand and the device’s maximum capacity, representative of realistic scenarios.  
In addition, we assume that the impact of model aggregation step on training time is negligible due to the server's abundant computational capacity. Assigning tasks in this manner has the following properties: First, the alignment score is highest for the job that utilizes the most resources on the device across all resource types. Second, if a device has ample availability of a particular resource, tasks that require more of that resource will receive higher scores compared to tasks with similar overall resource requirements. Third, only jobs that can be fully satisfied by the available resources on a device are considered, thereby preventing any risk of oversubscription.

\SetKwInput{KwInput}{Input}
\SetKwInput{KwOutputOne}{Output} 
\SetKwInput{KwOutputTwo}{Server Initializes}
\SetKwInput{KwOutputThree}{Server Executes}  
\SetKwInput{KwOutputFour}{ClientUpdate} 
\SetAlgoNoEnd
\SetAlgoNoLine
\begin{algorithm}[t]

\DontPrintSemicolon
  \KwInput{\\
  \hspace{2em}$\mathcal{V}_0$: Set of currently occupied devices; \\
  \hspace{2em}${s_{k,m}}$: Frequency of device $k$ scheduled to\\
  \hspace{4.1em}Job $m$;\\
  \hspace{2em}$R_m$: Maximum round for Job $m$; \\
  \hspace{2em}$l_m$: Desired loss for Job $m$.}
  \KwOutputOne{\\
  \hspace{2em}$\mathcal{V}_m = \{\mathcal{V}_m^{*1}, ..., \mathcal{V}_m^{*R_m}\}$: Scheduling plans, each\\
  \hspace{2em}with $C_m$ devices for Job $m$;\\
                }
  \KwOutputTwo{Initialize \textit{r} = 0, $\mathcal{V}_0 = \emptyset$, $w_{0,m}^g$;\\
                }
  \KwOutputThree{\\
    \For{$r = \{1, \cdots, R_m\}$ and $l_m$ is not achieved}{
        Server collects resource availability from devices;\\
        Server evaluates $\mathcal{R}_{k,m}^r$ for Job $m$ and each Device $k$;\;
        Server computes $\mathcal{F}_{k,m}^r$ by using Formula~\ref{formula5};\\
        Server calculates $Score_{k,m}^r$ by using Formula~\ref{formula3};\\
        Select top-$C_m$ devices for Job $m$: \;
        \quad $Score_{k,m}^r$: $\mathcal{V}_m^{*r} \leftarrow \text{Top-}C_m(Score_{k,m}^r)$;\\
        Update frequency vector $\boldsymbol{s}_m=\{s
        _{1,m},\cdots,s_{K,m}\}$ and occupied device set $\mathcal{V}_0$;\;
       
        {
        
         Server aggregates parameters:\;
        \quad $w_{r+1,m}^g \leftarrow \sum_{k \in \mathcal{V}_m^{*r}} p_k^r w_{r,m}^k$;\;
        
        Server sends $w_{r+1,m}^g$ to clients in $\mathcal{V}_m^{*(r+1)}$;
        }
        }
        \textbf{end}\;
        }

  \KwOutputFour{\\
  
  \qquad \For{each device $k \in \mathcal{V}_m^{*r}$ in parallel}{
  \qquad Client \textit{k} receives model parameter $w_{r,m}^g$;\;
  \qquad $w^k\leftarrow w_{r,m}^g$;\;
  \qquad \For{each local epoch $e = 1, \cdots, E$}{
  \qquad \For{each batch $b$ in $B_k$}{
        \qquad$w^k \leftarrow w^k - \eta \nabla f(w^k; b)$\;
    }
    \qquad \textbf{end}\;
    }
    \qquad \textbf{end}\;
    }
  \qquad $w_{r,m}^k \leftarrow w^k$\;
  \qquad Upload $w_{r,m}^k$ to the server;\; 
  \qquad \textbf{end}\;
  
  }

\caption{FedACT Algorithm}
\label{algo1}

\end{algorithm}
\setlength{\textfloatsep}{5pt}

{\bf Participation Fairness.}
Our preliminary experiments indicate that focusing solely on resource alignment does not yield optimal outcomes, particularly in scenarios involving heterogeneous data distribution. To mitigate the impact of data heterogeneity, we incorporate participation fairness into the calculation of the alignment score. Participation fairness ensures equitable involvement of devices in the FL training process, which can be measured by the standard deviation of the time each device is assigned to a job \cite{pitoura2006load, finkelstein2008fairness}. To avoid overburdening some devices while leaving others idle, it is important to fairly distribute training jobs. We next explore how the alignment scoring mechanism guarantees that all devices are allocated a fair share of the workload, promoting fairness without compromising training efficiency.

\setlength{\belowcaptionskip}{-1pt} 
\begin{table*}
\centering

\caption{A summary of the datasets and models used in our experiments. The dataset size indicates the number of training samples and test samples (number of training samples/number of test samples). Emnist (L) and Emnist (D) represent Emnist-Letters and Emnist-Digits, respectively.}
\begin{tabular}{lcccccc}
\toprule
& \multicolumn{3}{c}{\textbf{Group A}} & \multicolumn{3}{c}{\textbf{Group B}} \\
\cmidrule(lr){2-4} \cmidrule(lr){5-7}
\textbf{Datasets} & \textbf{Cifar-10} & \textbf{Emnist (L)} & \textbf{Emnist (D)} & \textbf{Fashion-Mnist} & \textbf{Cifar-10} & \textbf{Mnist} \\
\midrule
Features & 32$\times$32 & 28$\times$28 & 28$\times$28 & 28$\times$28 & 32$\times$32 & 28$\times$28 \\
Model Architecture & VGG-16 & CNN & LeNet-5 & CNN & ResNet-18& AlexNet \\
Model Parameters & 26,233K & 3,785K & 62K & 225K & 598k & 3,275K \\
Dataset Size & 50k/10k & 124.8k/20.8k & 240k/40k & 60K/10K & 50K/10K & 60K/10K \\
Local epochs & 5 & 5 & 5 & 5 & 5& 5 \\
Mini-batch size & 30 & 10 & 64 & 10 & 30 & 64 \\
\bottomrule
\end{tabular}
\label{dataset}
\end{table*}

\begin{table*}
\caption{Elapsed wall-clock time needed to reach the convergence accuracy and target accuracy (shown in parentheses) for different methods in Group A. ``/'' denotes that the method failed to achieve the target accuracy.}
\centering
\small 
\setlength{\tabcolsep}{4pt}
\begin{tabular}{cccccccccccc}
\hline
& \multicolumn{5}{c}{Convergence Accuracy} && \multicolumn{5}{c}{Time (min)} \\
\cline{2-6} \cline{8-12}
& Random & Greedy & Genetic & MJ-FL & FedACT && Random & Greedy & Genetic & MJFL & FedACT\\
\hline
& \multicolumn{11}{c}{IID} \\
\hline
LeNet & 0.9924 & 0.990 & 0.9931 & 0.9937 & \textbf{0.9943} & LeNet(0.99) & 15.13 & 17.25 & 6.42 & 5.18 & \textbf{3.92}\\
CNN & \textbf{0.942} & 0.926 & 0.928 & 0.935 & \textbf{0.942} & CNN(0.93) & 58.41 & 30.17 & 109.68 & 19.18 & \textbf{11.52} \\
VGG & \textbf{0.611} & 0.52 & 0.557 & 0.602 & 0.605 & VGG(0.55) & 134.3 & / & 229.7 & 59.10 & \textbf{43.25}\\
\hline
& \multicolumn{11}{c}{Non-IID} \\
\hline
LeNet & 0.990 & 0.987 & 0.986 & 0.990 & \textbf{0.991} & LeNet(0.984) & 42.56 & 42.73 & 31.28 & 21.87 & \textbf{15.02}\\
CNN & \textbf{0.90} & 0.84 & 0.80 & 0.887 & \textbf{0.90} & CNN(0.80) & 43.61 & 97.33 & 28.08 & 20.73 & \textbf{14.82} \\
VGG & 0.55 & 0.44 & 0.54 & 0.56 & \textbf{0.576} & VGG(0.55) & 2471.4 & / & 1168.5 & 464.7 & \textbf{286.4}\\
\hline
\end{tabular}
\label{table3}
\vspace{-13pt}
\end{table*}

The participation fairness of device \textit{k} for job \textit{m} in round \textit{r} is evaluated based on the variation in the scheduling frequency of each device for the job, represented as follows:
\begin{equation}
\mathcal{F}^r_{k,m}(\mathcal{V}^r_m) = 1 - \left(s_{k,m}^r - \frac{1}{|\mathcal{K}|} \sum_{k \in \mathcal{K}} s_{k,m}^r\right)^2
\label{formula5}
\end{equation}
Here, $s^r_{k,m}$ denotes the participation count of device \textit{k} to be assigned to job \textit{m}. It is determined by counting the number of times device \textit{k} is scheduled to participate in job \textit{m} across all scheduling plans for the job, i.e., $\{ \mathcal{V}^1_m,...,\mathcal{V}^r_m\}$. 
At the start of training (round 0), each $s^0_{k,m}$ is initialized to 0. Based on its meaning of participation count, $s^{r+1}_{k,m}$ can be expressed as follows:
\begin{equation}
s^{r+1}_{k,m} = 
\begin{cases}
s^r_{k,m} + 1, & \text{if device } k \in \mathcal{V}^r_m \\
s^r_{k,m}, & \text{otherwise}
\end{cases}
\end{equation}

It is worth noting that the participation fairness inspired from \cite{petrangeli2014multi} is measured by the deviation of the frequency of each device assigned to a job, which differs from the fairness concept (the bias of models toward certain features) in machine learning \cite{mehrabi2021survey}. 
It ensures that devices are selected in a balanced manner, increasing participating chances for underutilized devices and mitigating the adverse impact on FL convergence and thus the learning outcome. Such benefits are especially crucial for multi-job FL across heterogeneous devices and datasets.

\section{Experiments}
\label{sec:experiment}

In this section, we present the experimental comparison of {\em FedACT} with four state-of-the-art approaches across five commonly used datasets to validate its effectiveness. We begin by detailing the experimental setup and then report the results of our evaluation.

\subsection{Experimental Setup}

\textbf{Implementation.} We implemented {\em FedACT} and baseline approaches using Python 3.9 and PyTorch 2.0 to simulate both the parameter server and clients. In all experiments, we assume the availability of 100 devices, with 10\% of the devices selected for each job in every communication round. We conducted the experiments on a server equipped with four NVIDIA RTX A4000 GPUs, an Intel(R) Core(TM) i9-10900X CPU, and 64GB of RAM. We used Stochastic Gradient Descent (SGD) as the optimizer and set the number of local training epochs to 5.

\textbf{Tasks, Datasets, and Models.} 
We consider various image classification tasks utilizing the CIFAR-10 \cite{krizhevsky2009learning}, MNIST \cite{lecun1998gradient}, EMNIST-Letters\cite{cohen2017emnist}, EMNIST-Digits\cite{cohen2017emnist}, and Fashion-MNIST \cite{xiao2017fashion} datasets during the training process. In the IID scenario, a predetermined number of images from the entire training set are randomly assigned to each device. To simulate the Non-IID setting, we first categorize the training data by class and partition each category into 20 subsets. Each device is then randomly assigned two categories and selects one subset from each category to form its local training set. In this experiment, we define three jobs to be executed simultaneously as part of two distinct groups. Group A includes LeNet-5 (LeNet) \cite{lecun1998gradient}, CNN (CNN-A-IID and CNN-A-Non-IID) \cite{lecun1998gradient}, and VGG-16 (VGG) \cite{simonyan2014very}, while Group B contains AlexNet \cite{krizhevsky2012imagenet}, CNN (CNN-B) \cite{lecun1998gradient}, and ResNet-18 (ResNet) \cite{he2016deep}, with each model training representing an individual job. The model complexities of both groups are ranked as follows: \( \text{LeNet} < \text{CNN (CNN-A-IID and CNN-A-Non-IID)} < \text{VGG and AlexNet} < \text{CNN-B} < \text{ResNet} \). 

\begin{table*}
\caption{Elapsed wall-clock time needed to reach the convergence accuracy and target accuracy (shown in parentheses) for different methods in Group B. ``/'' denotes that the method failed to achieve the target accuracy.}
\centering
\small 
\setlength{\tabcolsep}{4pt}
\begin{tabular}{cccccccccccc}
\hline
& \multicolumn{5}{c}{Convergence Accuracy} && \multicolumn{5}{c}{Time (min)} \\
\cline{2-6} \cline{8-12}
& Random & Greedy & Genetic & MJ-FL & FedACT && Random & Greedy & Genetic & MJFL & FedACT\\
\hline
& \multicolumn{11}{c}{IID} \\
\hline
AlexNet & 0.9937 & 0.9834 & 0.9938 & 0.9939 & \textbf{0.9942} & AlexNet(0.9933) & 37.02 & / & 20.28 & 21.90 & \textbf{13.58}\\
CNN & 0.868 & 0.868 & 0.867 & 0.868 & \textbf{0.871} & CNN(0.867) & 123.10 & 36.88 & 40.08 & 31.35 & \textbf{21.08}\\
ResNet & 0.786 & 0.742 & 0.753 & 0.784 & \textbf{0.796} & ResNet(0.74) & 64.66 & 52.68 & 31.26 & 16.65 & \textbf{11.42} \\
\hline
& \multicolumn{11}{c}{Non-IID} \\
\hline
AlexNet & 0.989 & 0.872 & 0.985 & \textbf{0.990} & \textbf{0.990} & AlexNet(0.976) & 141.50 & 179.9 & 60.72 & 59.62 & \textbf{48.37}\\
CNN & 0.823 & 0.764 & 0.768 & 0.830 & \textbf{0.843} & CNN(0.73) & 46.82 & 71.38 & 22.95 & 16.51 & \textbf{12.83} \\
ResNet & 0.547 & 0.404 & 0.487 & 0.568 & \textbf{0.584} & ResNet(0.50) & 851.7 & / & 623.5 & 218.4 & \textbf{166.8}\\
\hline
\end{tabular}
\label{table4}
\vspace{2pt}
\end{table*}

\begin{table*}
\centering
\caption{Elapsed wall-clock time needed to reach the target accuracy in Single-Job FL using FedAvg. ``*'' denotes that it fails to reach the target accuracy.} 
\setlength{\tabcolsep}{7pt}
\begin{tabular}{ccccccc}
\toprule
\multicolumn{1}{c}{} & \multicolumn{3}{c}{IID/Non-IID} & \multicolumn{3}{c}{IID/Non-IID} \\
\cmidrule(lr){2-4} \cmidrule(lr){5-7}
Job & VGG & CNN & LeNet & ResNet & CNN & AlexNet \\
\midrule
Target Accuracy & 0.55/0.55 & 0.93/0.80 & 0.99/0.984 & 0.74/0.50 & 0.867/0.73 & 0.9933/0.976 \\
Time (min) & 140.7/2488.3 & 49.6/52.4 & 19.1/49.5 & */896.7 & 32.1/36.8 & 67.02/116.5 \\
\bottomrule
\end{tabular}
\label{table5}
\vspace{-13pt}
\end{table*}

CNN-A-IID consists of two convolutional (conv) layers with a kernel size of $3\times3$, the first having 32 channels and the second 64 channels. Each conv layer is followed by batch normalization and a $2\times2$ max pooling layer. The network also includes a flattening layer and three fully connected (fc) layers with 1568, 784, and 26 units, respectively. Additionally, we slightly modify CNN-A-IID to CNN-A-Non-IID, due to the poor convergence of CNN in the non-IID scenario of Group A. CNN-A-non-IID comprises of three $3\times3$ conv layers with 32, 64, and 64 channels, each using ReLU activation. The first two conv layers are followed by $2\times2$ max pooling and each conv layer is followed by a dropout layer with a rate of 0.05. The network also includes a flattening layer and two fc layers with 64 and 26 units, respectively. Additional experimental parameters are summarized in Table \ref{dataset}.

\textbf{Baselines Methods.} To comprehensively evaluate the performance of {\em FedACT}, we compare it against the following state-of-the-art baseline approaches.

\begin{enumerate}
    \item \textbf{Random}\cite{mcmahan2017communication}: Originally designed for single-job FL, we extend this method to multi-job FL by randomly selecting devices for each job in every communication round.
    \item \textbf{Greedy} \cite{wang2020cellular}: This baseline iteratively selects devices by maximizing a combined score of resource alignment and a participation-based penalty.
    \item \textbf{Genetic}\cite{barika2019scheduling}: It leverages heuristic initialization and evolutionary strategies to optimize device selection, aiming to maximize resource alignment and improve training efficiency.
    \item \textbf{MJ-FL}\cite{zhou2022efficient}: This multi-job FL baseline applies Bayesian Optimization to assign devices to jobs while considering both communication and computation capabilities.
\end{enumerate}

\textbf{Evaluation Metrics.} We assess the wall-clock training time required for each job to reach its specified target accuracy or complete a predefined number of communication rounds. Furthermore, we evaluate the average job completion time (JCT) across all job groups to measure the overall scheduling efficiency.

\subsection{Results and Analysis}

We first compare the performance of {\em FedACT} to the vanilla FL baseline under naive multi-job execution practice. We then evaluate it against several state-of-the-art multi-job FL scheduling approaches under both IID and Non-IID settings.

\begin{figure*}
\centering
\captionsetup[subfigure]{font=footnotesize, margin=0.16\linewidth}
\subfloat[]{%
    \includegraphics[height=4cm, width=0.30\textwidth]{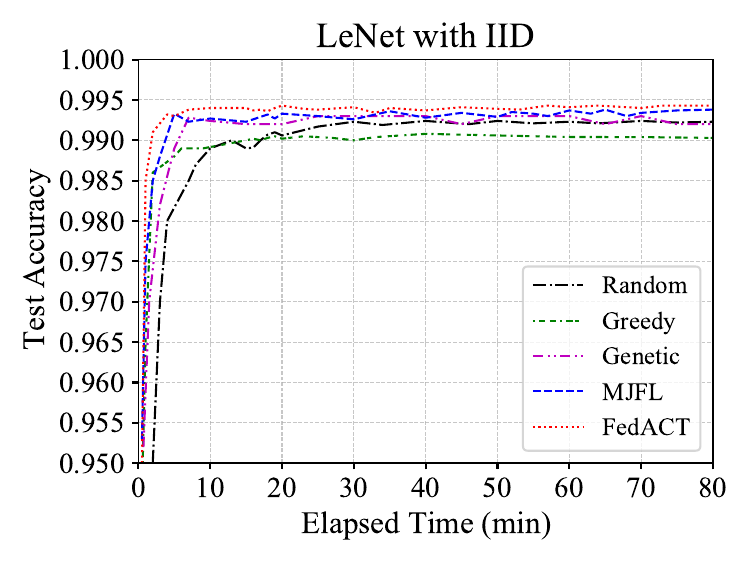}%
    \label{fig:subfig:lenet_iid_A}}
\hspace{0.01\textwidth}
\subfloat[]{%
    \includegraphics[height=4cm, width=0.30\textwidth]{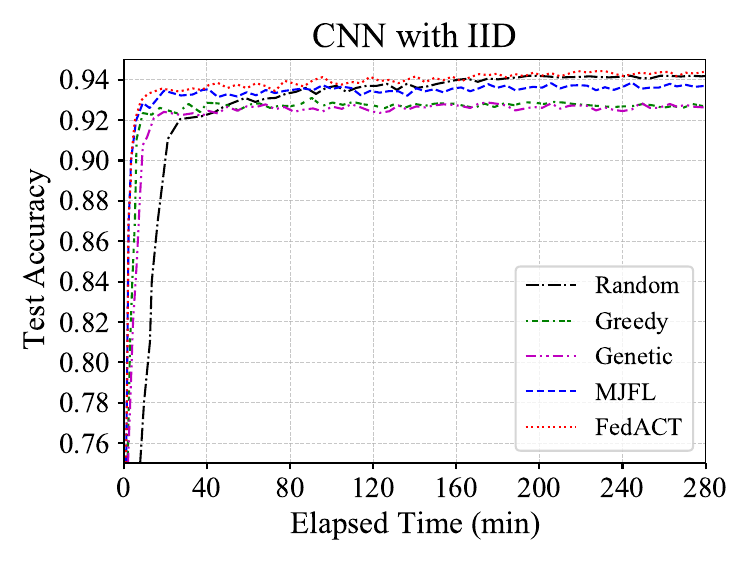}%
    \label{fig:subfig:cnn_iid_A}}
\hspace{0.01\textwidth}
\subfloat[]{%
    \includegraphics[height=4cm, width=0.30\textwidth]{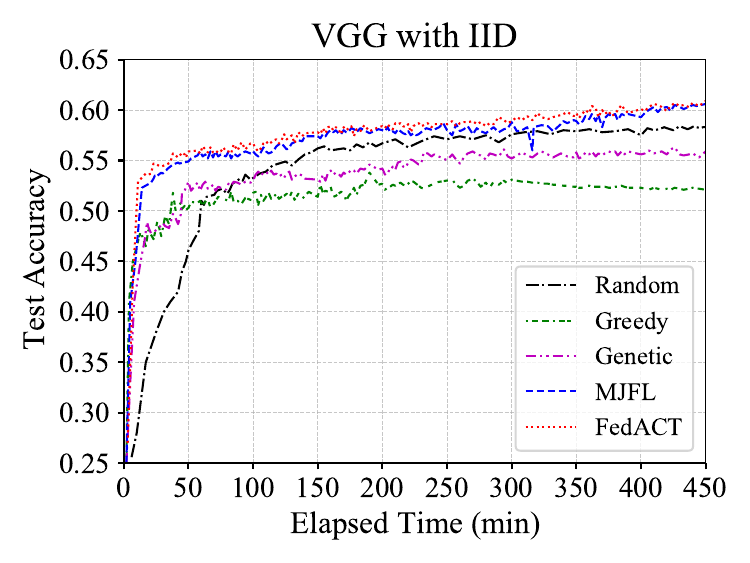}%
    \label{fig:subfig:vgg_iid_A}}
\caption{Test accuracy versus elapsed wall-clock time for different jobs in Group A with the IID distribution.}
\label{figure4}
\vspace{-20pt}
\end{figure*}
\begin{figure*}
\centering
\captionsetup[subfigure]{font=footnotesize, margin=0.16\linewidth}
\subfloat[]{%
   \includegraphics[height=4cm, width=0.30\textwidth]{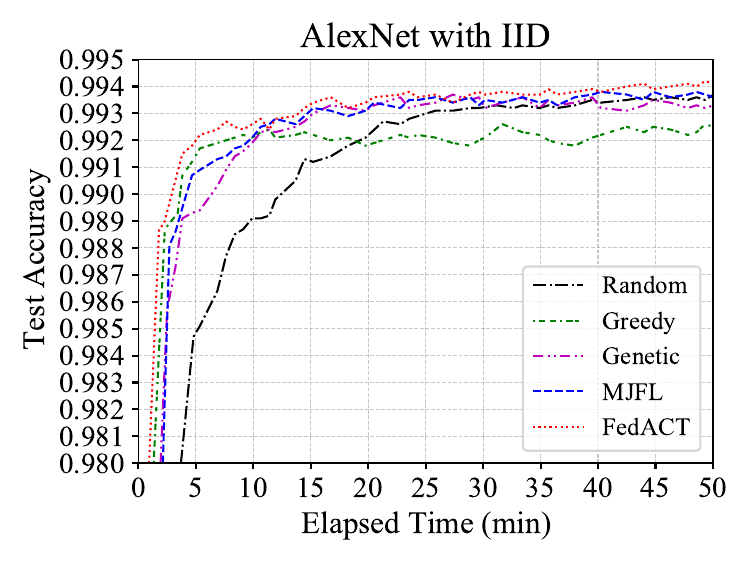}%
   \label{fig:subfig:alexnet_iid_B1}}
\hspace{0.01\textwidth}
\subfloat[]{%
   \includegraphics[height=4cm, width=0.30\textwidth]{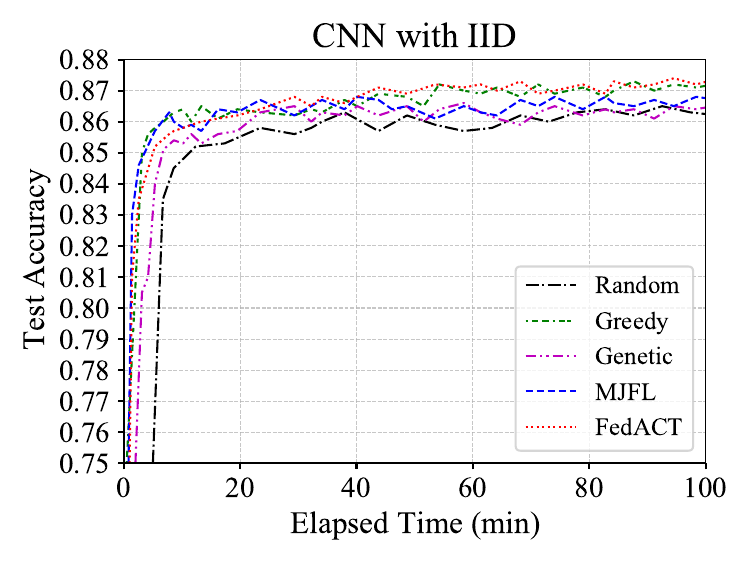}%
   \label{fig:subfig:cnn_iid_B}}
\hspace{0.01\textwidth}
\subfloat[]{%
   \includegraphics[height=4cm, width=0.30\textwidth]{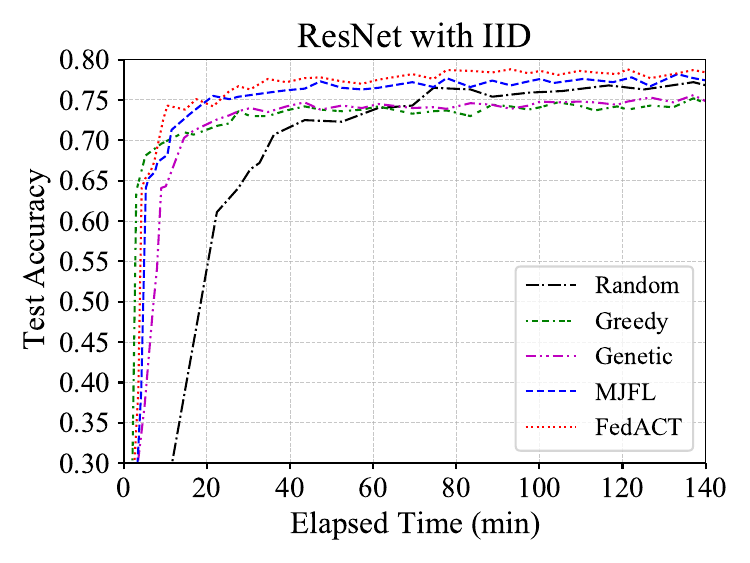}%
   \label{fig:subfig:resnet_iid_B}}
\caption{Test accuracy versus elapsed wall-clock time for different jobs in Group B with the IID distribution.}

\label{figure5}
\vspace{-20pt}
\end{figure*}

\subsubsection{Performance comparison against vanilla FL for multiple jobs}
To compare with the vanilla FL baseline under the same set of jobs, we sequentially execute each group of jobs using FedAvg, referred to as {\em Single-Job FL}. This approach is denoted as {\em Random} when adapting it to the multi-job setting. Table~\ref{table5} shows the wall-clock training time required for {\em Single-Job FL} to reach the target accuracy for each job, under both IID and Non-IID data distributions. As shown in Tables~\ref{table3},~\ref{table4}, and~\ref{table5}, {\em FedACT} significantly outperforms {\em Single-Job FL}, achieving up to 8.7\(\times\) faster wall-clock training time to reach a similar accuracy of 55\% for VGG on Non-IID data. Furthermore, {\em FedACT} consistently reaches the target accuracy in less time under both IID and Non-IID scenarios across both job groups. Specifically, for Group B jobs under Non-IID settings, {\em FedACT} achieves up to 5.4\(\times\), 2.8\(\times\), and 2.4\(\times\) speedup for ResNet (50\%), CNN (73\%), and AlexNet (98\%), respectively. Similarly, for Group A jobs under IID settings, it achieves speedups of up to 3.2\(\times\), 4.3\(\times\), and 4.9\(\times\) for VGG (55\%), CNN (93\%), and LeNet (99\%), respectively. 
In addition, even the Multi-Job FL framework using the Random method (Tables~\ref{table3},~\ref{table4}) achieves substantially better performance than {\em Single-Job FL}, further highlighting the importance of simultaneously training multiple FL jobs.

\begin{figure*}
\centering
\captionsetup[subfigure]{font=footnotesize, margin=0.16\linewidth}
\subfloat[]{%
   \includegraphics[height=4cm, width=0.30\textwidth]{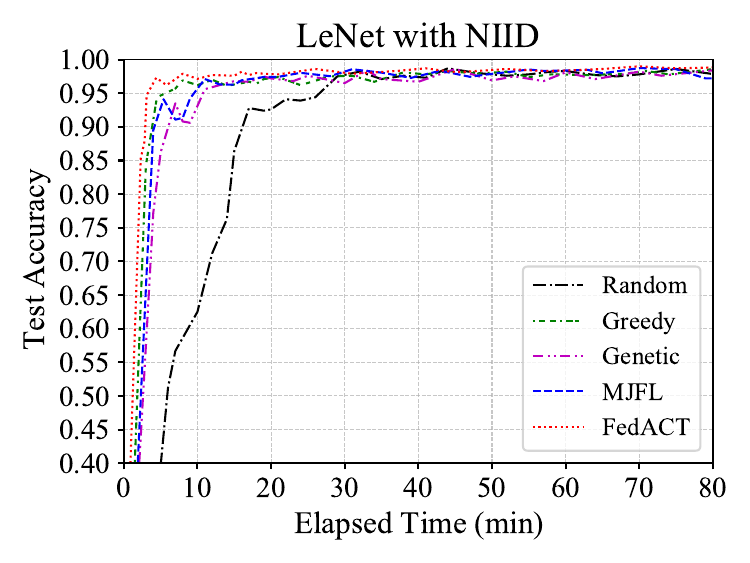}%
   \label{fig:subfig:lenet_iid3}}
\hspace{0.01\textwidth}
\subfloat[]{%
   \includegraphics[height=4cm, width=0.30\textwidth]{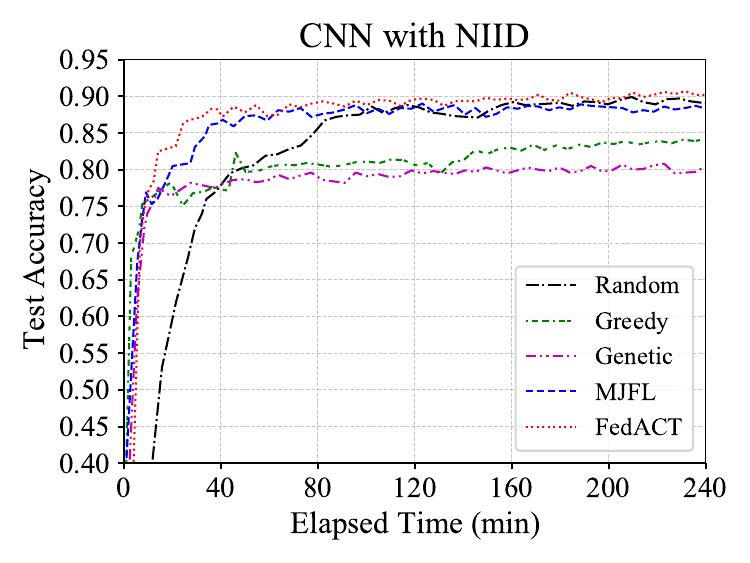}%
   \label{fig:subfig:cnn_iid3}}
\hspace{0.01\textwidth}
\subfloat[]{%
   \includegraphics[height=4cm, width=0.30\textwidth]{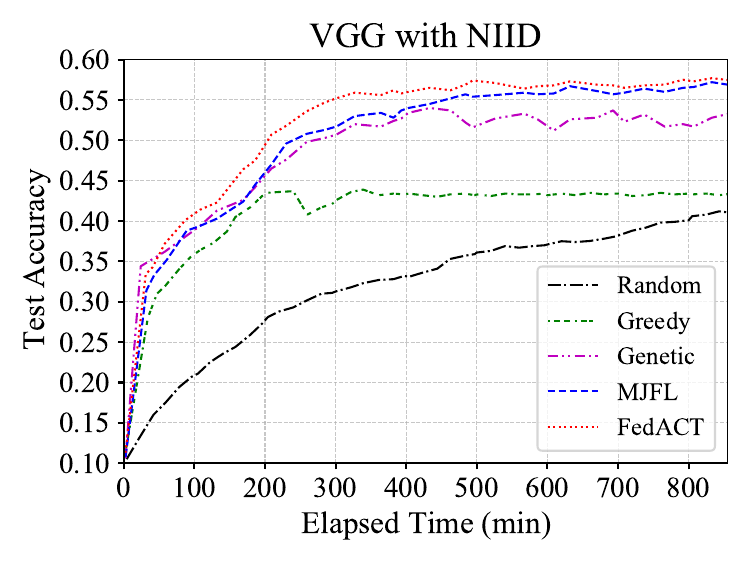}%
   \label{fig:subfig:vgg_iid3}}
\caption{Test accuracy versus elapsed wall-clock time for different jobs in Group A with the Non-IID distribution.}
\vspace{-13pt}
\label{figure6}

\end{figure*}

\subsubsection{Performance comparison against multi-job FL baselines}
In this section, we analyze the experimental outcomes of {\em FedACT} in comparison with existing multi-job FL scheduling methods under both the IID and the Non-IID settings.

\textbf{Results analysis under the IID setting:}
Figs.~\ref{figure4} and~\ref{figure5} show the elapsed wall-clock training time required to achieve convergence accuracy across all jobs, clearly demonstrating that our proposed method converges significantly faster than the baselines. Moreover, {\em FedACT} attains higher accuracy in a shorter training time compared to baseline approaches for both simple jobs (AlexNet in Fig.~\ref{figure4}(a)) and complex jobs (ResNet in Fig.~\ref{figure4}(c)). Tables~\ref{table3} and~\ref{table4}  demonstrate that all approaches achieve high convergence accuracy when trained on IID data. Specifically, for VGG in the IID scenario, {\em FedACT} achieves a final accuracy up to 16.35\% higher than Greedy, significantly outperforming all other methods. In addition, our proposed method achieves the target accuracy with significantly shorter training times compared to baseline methods, both in terms of individual job training time (up to 8.7\(\times\) faster) and the overall time for the whole training process (up to 6.26\(\times\) faster). In particular, focusing on CIFAR-10 with the VGG model, {\em FedACT} outperforms all baseline methods (by up to 3.1\(\times\) for Random, 3.1\(\times\) for Genetic, 1.4\(\times\) for MJFL) to reach the target accuracy of 55\%.

\begin{figure*}
\centering
\captionsetup[subfigure]{font=footnotesize, margin=0.16\linewidth}
\subfloat[]{%
   \includegraphics[height=4cm, width=0.30\textwidth]{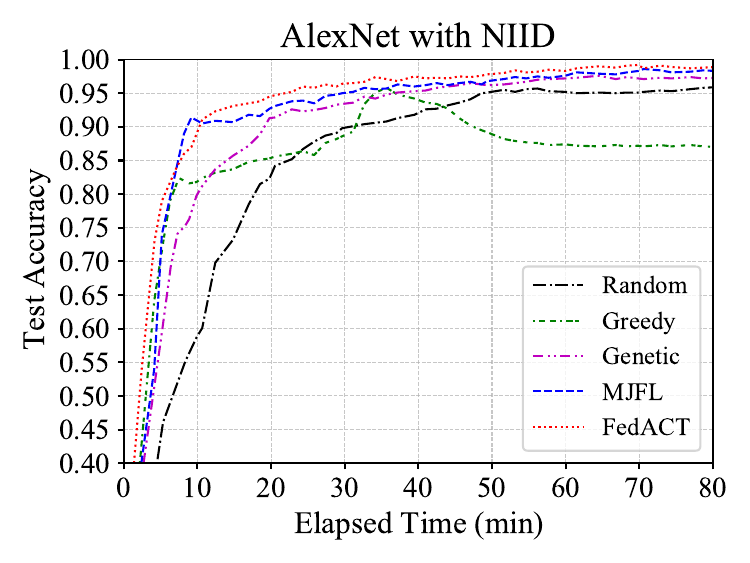}%
   \label{fig:subfig:lenet_iid4}}
\hspace{0.01\textwidth}
\subfloat[]{%
   \includegraphics[height=4cm, width=0.30\textwidth]{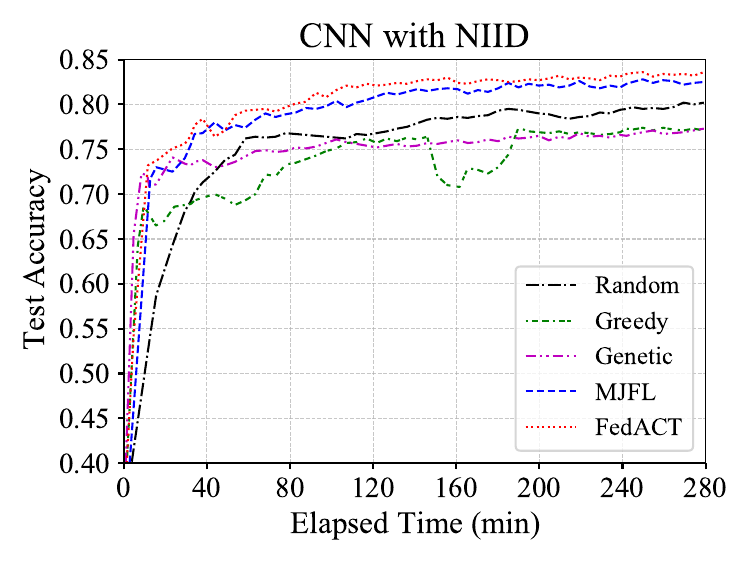}%
   \label{fig:subfig:cnn_iid4}}
\hspace{0.01\textwidth}
\subfloat[]{%
   \includegraphics[height=4cm, width=0.30\textwidth]{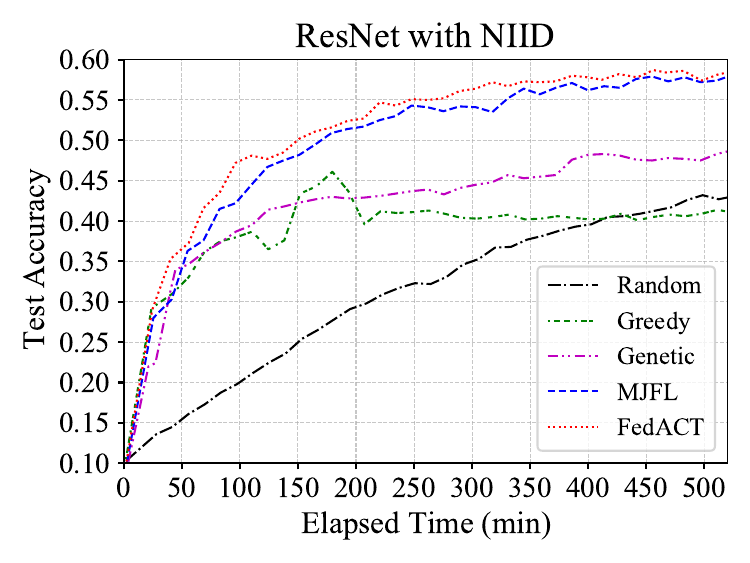}%
   \label{fig:subfig:vgg_iid4}}
\caption{Test accuracy versus elapsed wall-clock time for different jobs in Group B with the Non-IID distribution.}
\label{figure7}
\vspace{-20pt}
\end{figure*}

\textbf{Results analysis under the Non-IID setting:}
When the data distribution on devices is of Non-IID, the participation fairness defined in Eq.~(\ref{formula5}) has a significant impact on the accuracy. Figs.~\ref{figure6} and~\ref{figure7} illustrate the wall-clock training time required for each baseline to achieve the convergence accuracy of jobs in both groups under the Non-IID scenario. {\em FedACT} converges significantly faster than the other methods and demonstrates consistent performance across both simple and complex jobs. Specifically, it shows a significant advantage for complex jobs (VGG in Fig.~\ref{figure6}(c)). As shown in Tables~\ref{table3} and~\ref{table4}, the final accuracy of {\em FedACT} significantly outperforms the baseline methods, with accuracy improvements of +44.5\%, +19.9\%, +6.76\%, and +2.82\% over Greedy, Genetic, Random, and MJFL, respectively. Compared to baseline methods, our proposed approach reaches the target accuracy in significantly less time, achieving up to 8.63\(\times\) and up to 8.09\(\times\) reduction in average job completion time across Groups A and B. In addition, we observe that {\em FedACT} delivers better performance in the presence of Non-IID data due to dynamically adjusting the impact of participation fairness to mitigate the imbalance of devices. MJFL appears to be the closest counterpart to our method across all datasets, while {\em FedACT} consistently achieves the shortest training time, as shown in Figs.~\ref{figure6} and~\ref{figure7}. The reason behind this fact is that MJFL schedules devices to jobs without accounting for the intrinsic heterogeneity of device resources, whereas {\em FedACT} considers both resource alignment and participation fairness, showcasing its effectiveness.

\begin{figure*}
\centering
\captionsetup[subfigure]{font=footnotesize, margin=0.16\linewidth}
\subfloat[]{%
   \includegraphics[height=3.7cm, width=0.29\textwidth]{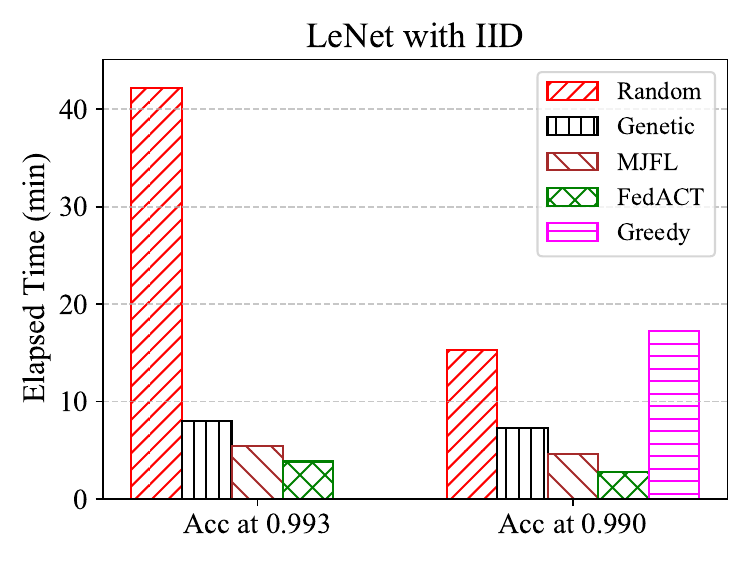}%
   \label{fig:subfig:lenet_iid5}}
\hspace{0.02\textwidth}
\subfloat[]{%
   \includegraphics[height=3.7cm, width=0.29\textwidth]{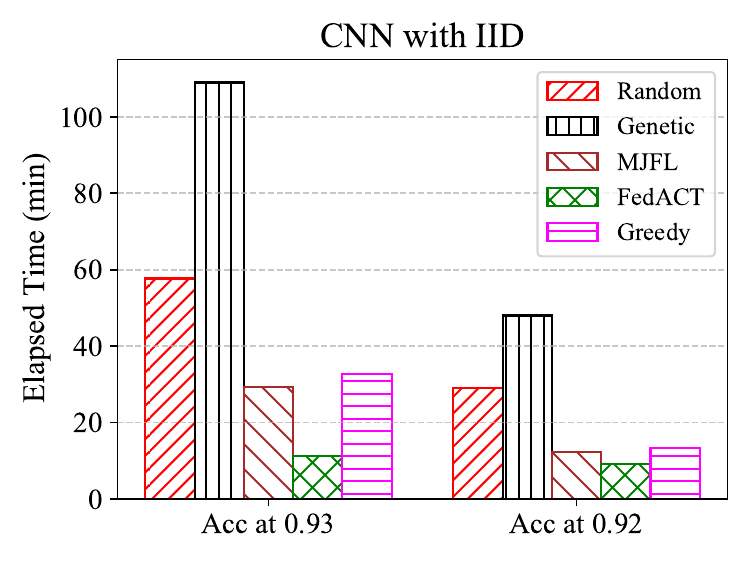}%
   \label{fig:subfig:cnn_iid5}}
\hspace{0.02\textwidth}
\subfloat[]{%
   \includegraphics[height=3.7cm, width=0.29\textwidth]{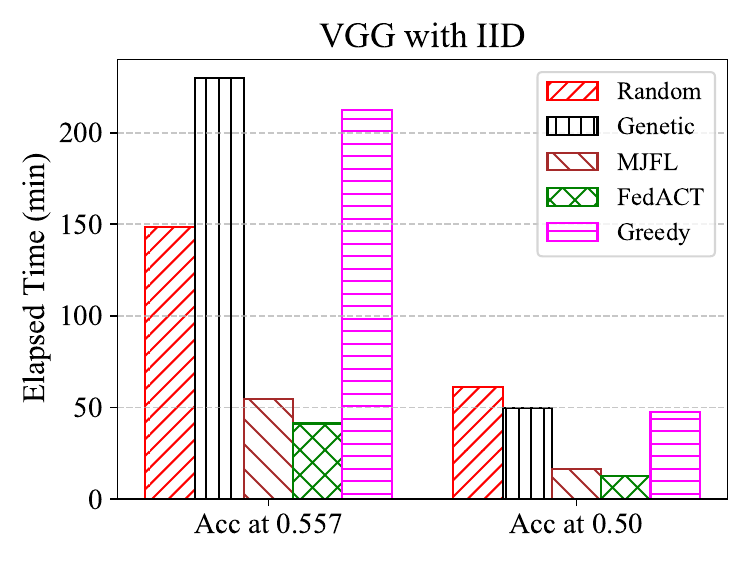}%
   \label{fig:subfig:vgg_iid5}}

\vspace{-10pt}
\subfloat[]{%
   \includegraphics[height=3.7cm, width=0.29\textwidth]{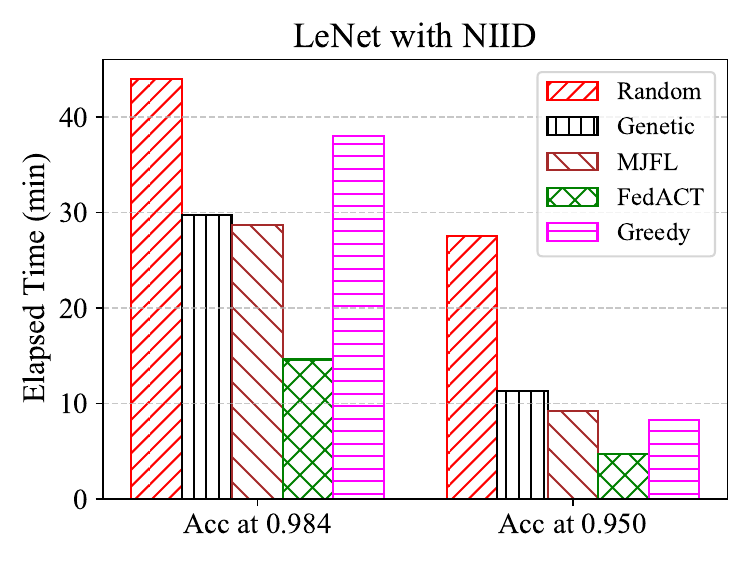}%
   \label{fig:subfig:lenet_niid5}}
\hspace{0.02\textwidth}
\subfloat[]{%
   \includegraphics[height=3.7cm, width=0.29\textwidth]{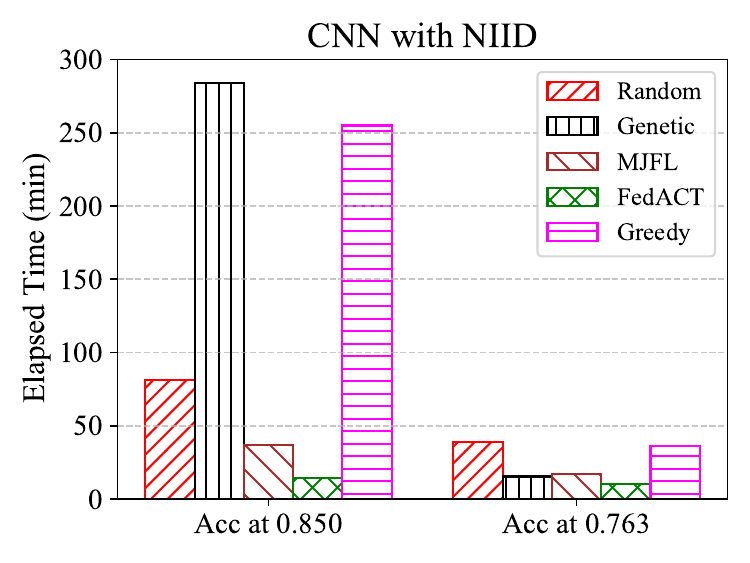}%
   \label{fig:subfig:cnn_niid5}}
\hspace{0.02\textwidth}
\subfloat[]{%
   \includegraphics[height=3.7cm, width=0.29\textwidth]{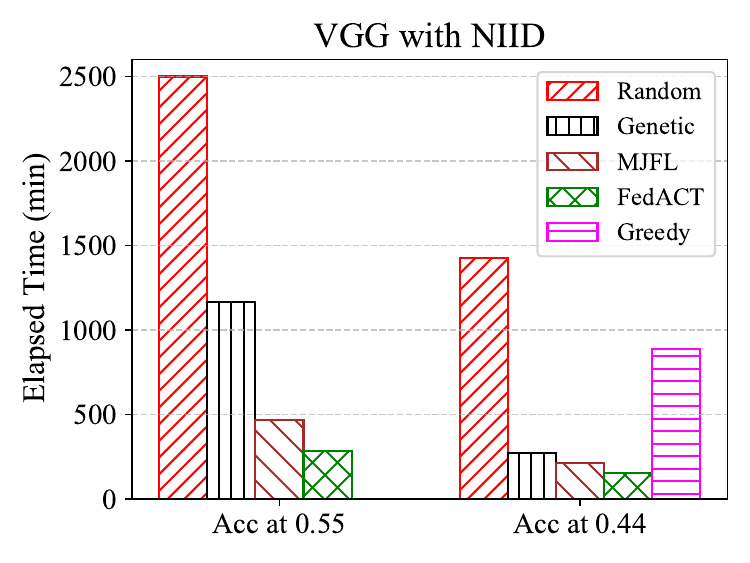}%
   \label{fig:subfig:vgg_niid5}}

\caption{Elapsed wall-clock training time required for each job of Group A to achieve the target convergence accuracy under IID and Non-IID data distributions.}
\vspace{-13pt}
\label{figure8}
\end{figure*}

\begin{figure*}[htbp]
\centering
\captionsetup[subfigure]{font=footnotesize, margin=0.16\linewidth}
\subfloat[]{%
   \includegraphics[height=3.7cm, width=0.29\textwidth]{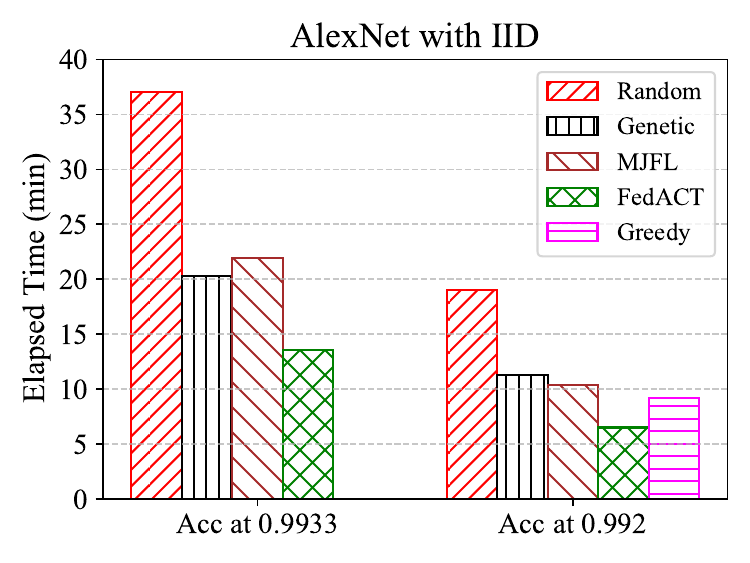}%
   \label{fig:subfig:lenet_iid6}}
\hspace{0.02\textwidth}
\subfloat[]{%
   \includegraphics[height=3.7cm, width=0.29\textwidth]{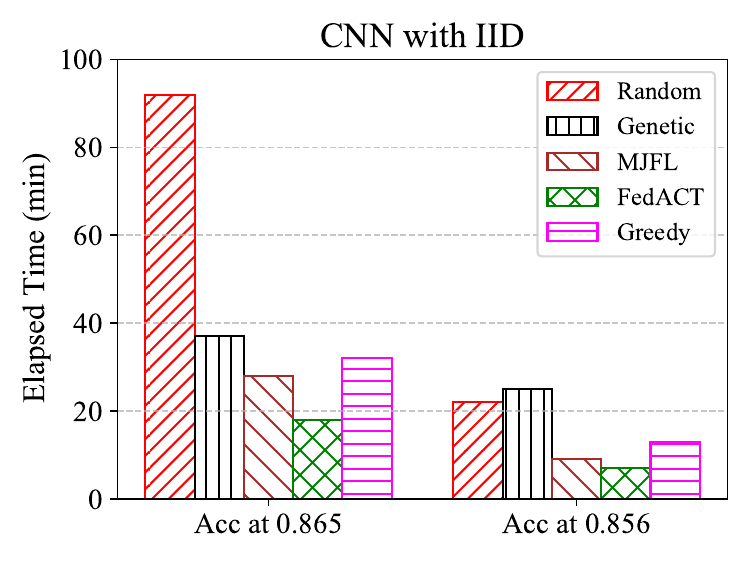}%
   \label{fig:subfig:cnn_iid6}}
\hspace{0.02\textwidth}
\subfloat[]{%
   \includegraphics[height=3.7cm, width=0.29\textwidth]{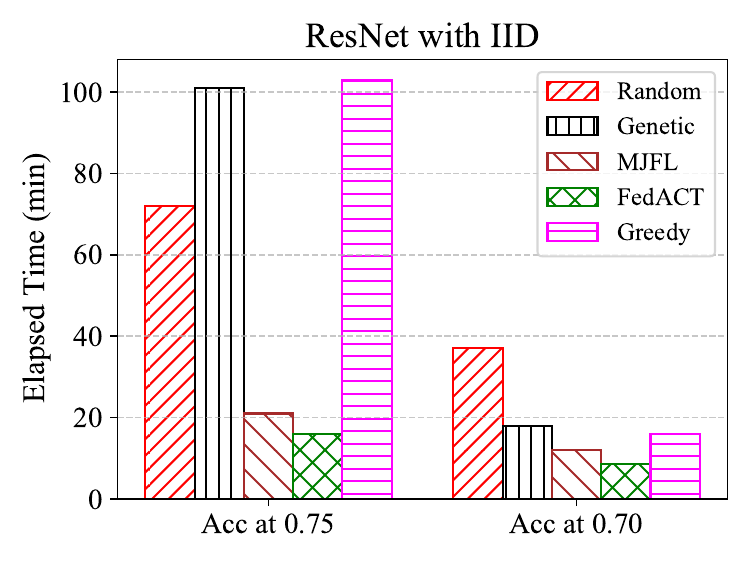}%
   \label{fig:subfig:vgg_iid6}}

\vspace{-10pt}
\subfloat[]{%
   \includegraphics[height=3.7cm, width=0.29\textwidth]{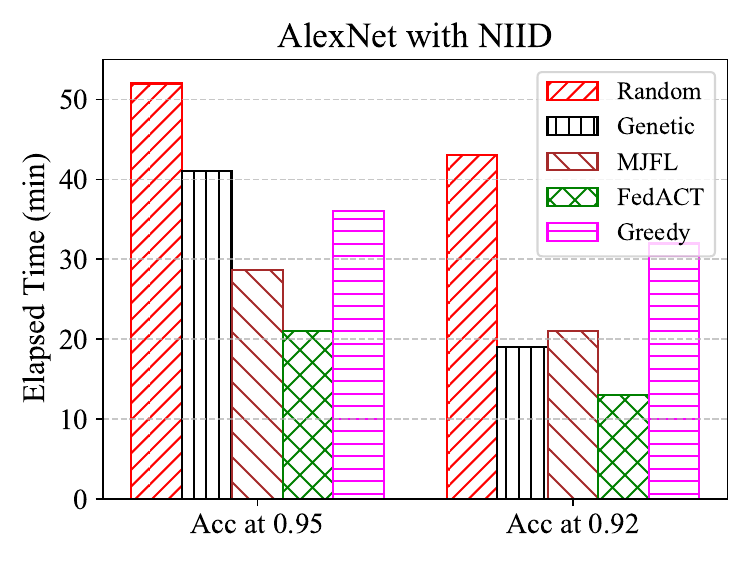}%
   \label{fig:subfig:lenet_niid6}}
\hspace{0.02\textwidth}
\subfloat[]{%
   \includegraphics[height=3.7cm, width=0.29\textwidth]{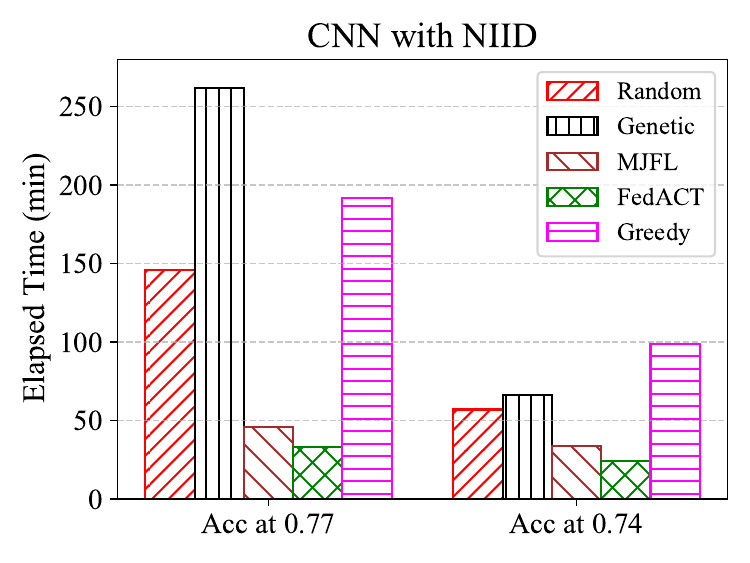}%
   \label{fig:subfig:cnn_niid6}}
\hspace{0.02\textwidth}
\subfloat[]{%
   \includegraphics[height=3.7cm, width=0.29\textwidth]{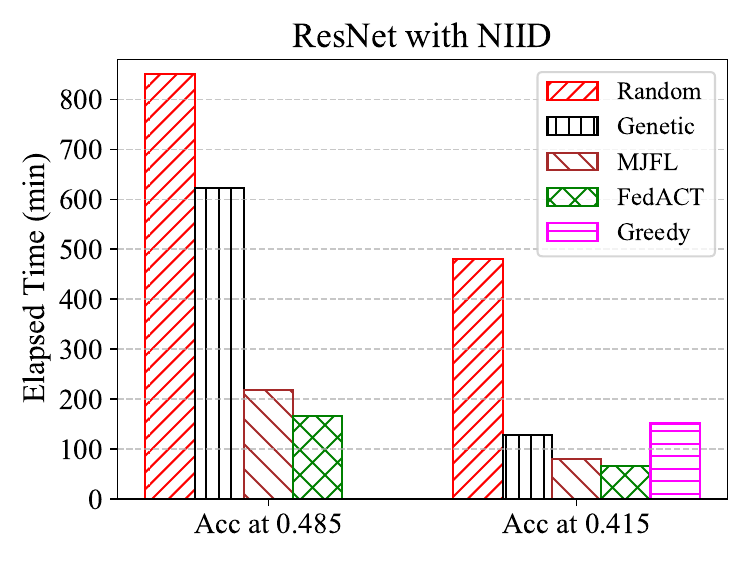}%
   \label{fig:subfig:vgg_niid6}}

\caption{Elapsed wall-clock training time required for each job of Group B to achieve the target convergence accuracy under IID and Non-IID data distributions.}
\vspace{-13pt}
\label{figure9}
\end{figure*}

\textbf{Model Accuracy vs. Job Completion Time.} We compare the job completion times required for each baseline method to reach different levels of target accuracy on both IID and Non-IID settings. As shown in Figs.~\ref{figure8} and~\ref{figure9}, {\em FedACT} achieves the best overall trade-off between model accuracy and training time compared to the baselines. The training time reduction is particularly notable for complex jobs (VGG in Fig.~\ref{figure8}(f) and ResNet in Fig.~\ref{figure9}(f)), especially with Non-IID data distributions. For instance, to achieve the target accuracy of 44\% on VGG, {\em FedACT} demonstrates a training time reduction of 9.3\(\times\), 5.8\(\times\), 1.8\(\times\), and 1.4\(\times\) compared to Random, Greedy, Genetic, and MJFL, respectively. Even for simple tasks, {\em FedACT} consistently yields the lowest training time across all six jobs. Specifically, for LeNet on Non-IID data, {\em FedACT} achieved up to 3.1\(\times\) faster training time than the baselines to reach the target accuracy of 98.4\%. Furthermore, {\em FedACT} accelerates the average JCT by up to 3.9\(\times\) and 8.3\(\times\) under IID and Non-IID settings, respectively. This efficiency improvement during training is attributed to {\em FedACT}'s resource-aligned device-to-job placement strategy. These results highlight {\em FedACT} as a promising solution for the simultaneous training of multiple jobs in FL with heterogeneous devices. 

\textbf{Impact of hyperparameters $\alpha$ and $\beta$.} The parameters $\alpha$ and $\beta$ control the relative influence of resource alignment and participation fairness, allowing {\em FedACT} to balance convergence speed and model accuracy under non-IID data distributions. In practice, we empirically set $\alpha$ and $\beta$ based on the information from previous execution and adjust them using small epochs. We increase $\alpha$ for fast convergence and increase $\beta$ mainly for high accuracy. {\em FedACT} exhibits stable behavior with the $\alpha$ and $\beta$ choices used in our implementation.

\textbf{Privacy and computation overhead.} {\em FedACT} requires each device to report resource availability and model parameters during training. Therefore, {\em FedACT} offers a similar level of privacy as state-of-the-art FL approaches by collecting the minimal information from devices. In addition, it can be further integrated with advanced privacy‐enhancing techniques such as homomorphic encryption \cite{paillier1999public}, secure aggregation \cite{bonawitz2016practical}, and differential privacy \cite{dwork2008differential}, to strengthen the protection of sensitive information. Furthermore, the additional computation and communication overheads in {\em FedACT} mainly arise from calculating alignment scores and generating scheduling plans, which are negligible considering its additional benefits.  

\section{Related Work}
\label{sec:relatedwork}
Existing related efforts (\cite{bonawitz2019towards,liu2020fedvision, lai2021oort,li2022pyramidfl,li2021talk,guo2024fedcore,jin2016enabling}, {\em etc.}) primarily focused on device scheduling in single-job FL systems. For instance, FedCore \cite{guo2024fedcore} addresses device heterogeneity by leveraging decentralized coresets to select representative local data, enabling efficient resource utilization and reducing the impact of straggler devices on model training. However, real-world applications have increasingly required the concurrent learning of multiple models in heterogeneous federated learning settings, posing critical open challenges for fast and efficient multi-job FL.
The varying computing capabilities and diverse data distributions across devices make it challenging to achieve a high-quality global model within a limited timeframe. Suboptimal device-to-job assignment prolongs the average job completion time, while the presence of data heterogeneity among devices further exacerbates the learning efficiency and negatively impacts the learning outcome with declined global model accuracy. 

Only a few recent studies \cite{liu2022multi, askin2024fedast, wei2022joint} have attempted to address the challenges of simultaneous training of multiple FL jobs. MJFL \cite{liu2022multi} selects devices using either Bayesian optimization or reinforcement learning to minimize training time. It incorporates communication and computation capabilities into its cost model, aiming to improve training efficiency and enhance global model accuracy. In contrast, {\em FedACT} explicitly considers device-to-job placement with awareness of resource heterogeneity and incorporates participation fairness, ensuring both efficient resource utilization and balanced contributions from participating devices across multiple concurrent jobs. \cite{bhuyan2022multi} formulates device assignment in FL as a bandit problem, utilizing local training losses as scoring metrics. It assigns devices either uniformly at random or in a round-robin manner. \cite{siew2023fair} proposes a biased device sampling strategy that prioritizes devices with higher local losses. However, their device selection strategies based solely on local training losses fail to capture the inherent heterogeneity among devices and jobs in real-world FL environments. 
In FEL \cite{wei2022joint}, the multi-job federated edge learning problem is formulated as a mixed-integer optimization task, with the objective of minimizing the overall system cost while satisfying resource and performance constraints. Several algorithms were proposed to decompose the original optimization task into sub-problems to be solved iteratively and independently. Venn \cite{liu2023venn} proposes a contention-aware job scheduling strategy combined with a resource-aware device-to-job matching mechanism to reduce the average job completion time (JCT) of concurrent FL jobs. Nevertheless, these approaches overlook fair participation of devices across jobs, which would degrade the overall learning outcome. FedAST \cite{askin2024fedast} introduces a buffered asynchronous federated training approach to mitigate the straggler issue and dynamically allocates device resources for heterogeneous jobs. Devices receive training requests from the server and execute them in a first-come-first-served manner, which may cause smaller jobs to wait for larger ones to finish, prolonging the average job completion time.


\section{Conclusion}
\label{sec:end}
This paper presents {\em FedACT}, a novel resource heterogeneity-aware device scheduling approach designed for the concurrent training of multiple federated learning jobs. {\em FedACT} enhances resource utilization and reduces average job completion time by efficiently assigning appropriate devices to each job based on an alignment scoring mechanism. This mechanism is composed of resource alignment between each job’s resource demands and the available resources on devices, and participation fairness. Participation fairness ensures balanced participation of devices across jobs, mitigating the impact of data heterogeneity. {\em FedACT} generates an optimal scheduling plan for each job by prioritizing devices with higher alignment scores. We carried out extensive experiments using five real-world models and five benchmark datasets under both IID and Non-IID distributions. Experimental results indicate that {\em FedACT} substantially outperforms both single-job FL and baseline scheduling methods, achieving up to 44.5\% higher accuracy, 8.6\(\times\) speedup for individual jobs, and 8.3\(\times\) faster average JCT. In future work, we plan to extend {\em FedACT} by incorporating asynchronous model updates, enabling devices to train multiple jobs concurrently without additional delays caused by server-side aggregation.

\bibliographystyle{IEEEtran}
\bibliography{sample-base}

\end{document}